\documentclass[runningheads]{llncs}
\usepackage[T1]{fontenc}
\usepackage[hidelinks]{hyperref}
\usepackage{cite}
\usepackage{booktabs}
\usepackage{graphicx}
\usepackage{amsmath}
\usepackage{subcaption}
\usepackage{orcidlink}
\usepackage{enumitem}
\setlist{nosep}
\usepackage{url}
\usepackage{algorithm}
\usepackage{algpseudocode}
\begin{document}
\title{Vortex: Multi-Modal Fusion System for Intelligent Video Retrieval}
%
%
\author{
Duc-Tho Nguyen\inst{1,2}\orcidlink{0009-0007-4595-486X}\thanks{Corresponding author} \and
Hieu-Hoc Tran-Minh\inst{1,2}\orcidlink{0009-0003-3787-6719} \and
Khanh-Hoa Lam\inst{1,2}\orcidlink{0009-0009-5014-5025} \and
Hoang-Nhut Ly\inst{1,2}\orcidlink{0009-0007-7153-1027} \and
Huu-Phuc Huynh\inst{1,2}\orcidlink{0009-0006-1170-7409} \and
 Thanh-Tien Tran\inst{1,2}\orcidlink{0009-0002-6077-2021} \and
 Trung-Nghia Le\inst{1,2}\orcidlink{0000-0002-7363-2610}
}
\authorrunning{D.-T. Nguyen et al.}
%
\institute{ University of Science, VNU-HCM, Ho Chi Minh City, Vietnam \and
Vietnam National University, Ho Chi Minh City, Vietnam}
\maketitle              
%

\begin{abstract}
This paper presents Vortex, the multimodal video retrieval system developed by our team, FocusOnFun, for the Ho Chi Minh City AI Challenge 2025, designed to advance intelligent multimedia search and temporal reasoning. The system integrates adaptive keyframe extraction, multimodal metadata generation from vision-language and speech models, and a hybrid retrieval strategy that fuses CLIP and SigLIP2 embeddings through Reciprocal Rank Fusion to balance global and fine-grained semantics. To enhance interactivity, Vortex incorporates Rocchio-based relevance feedback and a multi-stage temporal search mechanism for sequential event alignment. Built on Milvus and Elasticsearch, the architecture enables scalable indexing and efficient retrieval. Evaluated in the official competition, our FocusOnFun team's system achieved a score of 79.6/88 (90.5\%) in the Preliminary Round and was further evaluated in the Final Round, achieving an `Excellent' overall performance with `Outstanding' results in the question-answering (QA) task. This demonstrating the complementary strengths of CLIP and SigLIP2 and confirming the effectiveness of the hybrid retrieval approach. The system establishes a robust foundation for future research in intelligent, context-aware, and interactive video retrieval.

\end{abstract}

\section{Introduction}

The Ho Chi Minh AI Challenge (AIC) \cite{AIChallenge2025} is a recurring scientific competition in Vietnam dedicated to advancing intelligent multimedia retrieval. Its design closely follows and aligns with leading international benchmarks, including the Video Browser Showdown (VBS) \cite{rossetto2025results2025videobrowser} and the Lifelog Search Challenge (LSC) \cite{tran2025stateoftheartlifelogretrievalreview}. The competition focuses on building intelligent assistant systems capable of performing deep semantic analysis and retrieval across large-scale multimedia databases.

The AIC'25 Final Round featured four distinct search tasks. The first is Textual Known-Item Search (Textual KIS), a core task in which systems must locate a specific video segment based on a natural language description. The second, introduced in the final round, is Video Known-Item Search (Video KIS), which requires finding a segment based on a short query video clip. The third task, Question Answering (Q\&A), moves beyond traditional retrieval by requiring systems to not only identify the relevant video segment but also comprehend its content and generate a precise textual answer. The fourth and most complex task, Temporal Retrieval and Alignment of Key Events (TRAKE), challenges systems to retrieve a video containing an entire sequence of described events and to accurately align each event with its corresponding semantic keyframe.

In this paper, we present Vortex, a comprehensive multi-modal video retrieval system, for the AIC'25. Designed as an end-to-end solution, Vortex integrates an adaptive keyframe extraction pipeline (AutoShot~\cite{zhu2023autoshotshortvideodataset} with $L_2$-norm filtering) to efficiently manage extensive video datasets by minimizing redundancy while preserving essential visual information. To enable deep content-based retrieval, the system generates rich multi-modal metadata. Qwen2.5-VL~\cite{bai2025qwen25vltechnicalreport} extracts textual cues from video frames through optical character recognition and captioning, while Whisper~\cite{radford2022robustspeechrecognitionlargescale} provides temporally aligned automatic speech recognition transcriptions. We propose a hybrid retrieval strategy that balances search breadth and precision by generating dual embeddings using CLIP~\cite{radford2021learningtransferablevisualmodels} for global semantic context and SigLIP2~\cite{tschannen2025siglip2multilingualvisionlanguage} for fine-grained detail recognition. Their results are combined through Reciprocal Rank Fusion (RRF)~\cite{Cormack2009ReciprocalRF} to produce an optimized ranking. To address the competition’s most challenging queries, Vortex also integrates advanced interactive components such as a multi-stage Temporal Search for sequential "before, main, and after" event alignment and an interactive Relevance Feedback loop based on the Rocchio algorithm~\cite{10.1145/3366750.3366755}, enabling users to iteratively refine their searches and achieve higher retrieval accuracy across all task categories.

We participated in all four tasks of AIC’25 as the FocusOnFun team, achieved a score of 79.6/88 (90.5\%) in the Preliminary Round and achieved \textit{excellent overall performance} in the Final Round. Our system demonstrated strong robustness, particularly in handling new and diverse query types. For the Video KIS task, our rich metadata pipeline performed exceptionally well. The system generated natural-language descriptions for each query video and leveraged high-quality OCR, ASR, and object-filtering modules to accurately locate the target content, resulting in excellent performance. We also attained very good results on the TRAKE task. Most notably, we achieved outstanding performance on the Q\&A task, underscoring the effectiveness of our proposed system.

Our main contributions are summarized as follows:
\begin{itemize}
\item We propose Vortex, an end-to-end multi-modal video retrieval framework designed to address all four tasks of the Ho Chi Minh City AI Challenge 2025, integrating efficient data processing, semantic understanding, and interactive retrieval within a unified system.

\item We develop a two-stage keyframe extraction pipeline using AutoShot and $L_2$-norm filtering to efficiently capture representative visual content.

\item We build a multi-modal metadata preprocessing pipeline employing Qwen2.5-VL for OCR and captioning, and Whisper for aligned audio transcriptions.

\item We introduce a hybrid retrieval strategy that fuses dual embeddings from CLIP and SigLIP2 via RRF for improved ranking accuracy.

\item We propose interactive modules, including multi-stage Temporal Search and Rocchio-based Relevance Feedback, to enhance complex and iterative queries.
\end{itemize}

\section{Related Work}

The field of interactive multimedia retrieval has advanced considerably through benchmarking initiatives such as the LSC and VBS, which evaluate systems for event retrieval from large multimodal datasets using textual, visual, and ad hoc queries \cite{AIChallenge2025,rossetto2025results2025videobrowser}. LSC focuses on personal multimodal archives, emphasizing real-time interactive search across Known-Item Search (KIS), Question Answering (QA), and ad hoc tasks \cite{10.1145/3729459}. Recent reviews highlight key trends, including embedding-based retrieval (e.g., CLIP, BLIP), Large Language Model (LLM) integration for query optimization, and improved user interfaces for temporal and collaborative search \cite{tran2025stateoftheartlifelogretrievalreview}. In LSC’25, systems such as U-Cker further enhanced retrieval through LLM-driven query refinement and temporal reasoning, improving performance on ambiguous lifelog queries \cite{10.1145/3729459.3748688}.

VBS targets large-scale video archives with tasks involving textual KIS, visual KIS, and QA \cite{rossetto2025results2025videobrowser}. The 2025 results demonstrate the effectiveness of hybrid vision-language models for contextual understanding, with top systems like ViFi leveraging SigLIP for precise multimodal matching \cite{10.1007/978-981-96-2074-6_46}. Other notable examples, such as VideoEase, integrated ASR, OCR, and temporal reranking for complex sequence retrieval \cite{10.1007/978-981-96-2074-6_44}, while VITRIVR by Heller et al. \cite{10.1007/978-3-030-67835-7_41} exemplified effective temporal query mechanisms.

Building on these foundations, the previous AIC'24 adopted similar principles, emphasizing Vietnamese-language data and LLM-assisted query expansion. Systems like NewsInsight2.0 focused on LLM-based query optimization and temporal algorithms \cite{10.1007/978-981-96-4291-5_28}. Notably, top-performing systems at AIC'24 introduced multimodal fusion, particularly through CLIP–BEiT-3 hybridization, along with shot-based reranking and LLM-driven query optimization to handle temporal reasoning and ambiguity \cite{10.1007/978-981-96-4291-5_14,10.1007/978-981-96-4291-5_17}.

Vortex extends these prior efforts by unifying key advances while introducing novel components. Instead of the established CLIP-BEiT-3 fusion, we propose a new hybrid retrieval strategy pairing CLIP with SigLIP2 \cite{tschannen2025siglip2multilingualvisionlanguage}. SigLIP2, a state-of-the-art model, offers exceptional fine-grained detail recognition and localization, providing a powerful complement to CLIP's global semantic understanding. Furthermore, Vortex integrates this novel embedding fusion with a complete interactive loop, combining a multi-stage Temporal Search (Section \ref{sec:temporal_search}) for sequential queries with a Rocchio-based Relevance Feedback mechanism (Section \ref{sec:relevance_feedback}). This combination allows our system to not only execute complex temporal and semantic searches but also to iteratively refine results based on direct user feedback, achieving robust and adaptive performance.

\section{Proposed System}

\subsection{System Architecture}
\label{sec:architecture}

\begin{figure}[t!]
\includegraphics[width=\textwidth]{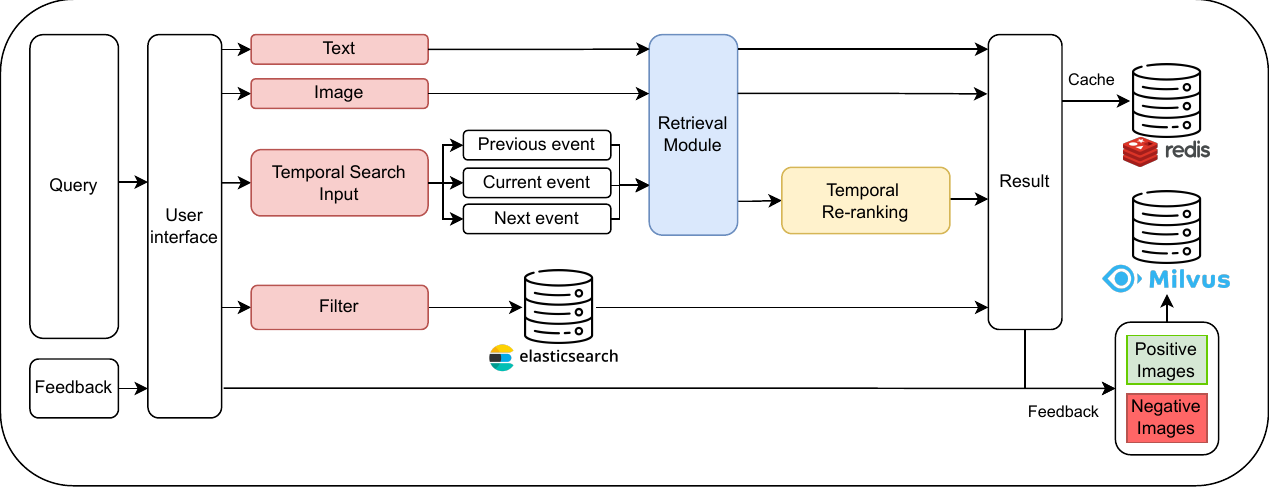}
\caption{Overall system architecture with two main workflows. (A) Query Processing supports text, image, temporal, and filter queries. Temporal queries include three text fields for previous, current, and next events, processed through the Retrieval Module and refined using the Temporal Re-ranking algorithm. Filter queries access scene descriptions or OCR metadata via Elasticsearch, and final results are cached in Redis for efficiency. (B) Feedback Loop allows users to label retrieved images as positive or negative, with feedback stored in Milvus to enhance retrieval accuracy.}
\label{fig:system_architecture}
\end{figure}

Our system is a comprehensive multimodal video retrieval platform designed for complex, large-scale search tasks. It supports both text and image queries, integrating hybrid vector search, metadata filtering, and advanced temporal and interactive retrieval. The overall architecture is shown in Fig.~\ref{fig:system_architecture}.

At its core, the system uses a dual-database backend: Milvus for high-dimensional vector search and Elasticsearch for text indexing and metadata filtering, with Redis providing low-latency caching. The workflow consists of two main phases. In the Preprocessing and Indexing phase, the system extracts keyframes (AutoShot with norm-based filtering), generates multimodal metadata (Qwen2.5-VL for OCR and captions, Whisper for ASR), and indexes CLIP and SigLIP2 embeddings. The Search and Interaction phase handles user queries through the Retrieval Module, which performs hybrid search with RRF, metadata filtering, temporal re-ranking, and Rocchio-based relevance feedback for efficient and adaptive retrieval.


\subsection{Data Pre-processing}

The overall data pre-processing pipeline is shown in Fig.~\ref{fig:preprocessing_pipeline}, including three main steps:

\begin{figure}[t!]
\centering
\includegraphics[width=\textwidth]{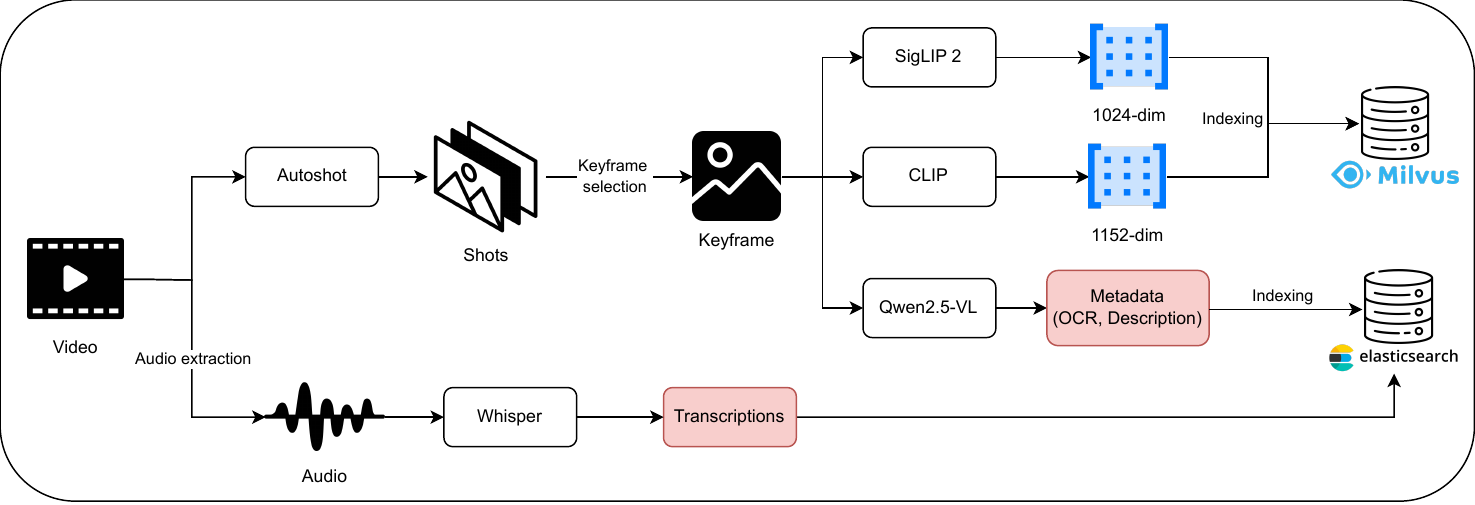}
\caption{Data pre-processing pipeline.} 
\label{fig:preprocessing_pipeline}
\end{figure}

\subsubsection{Keyframe Extraction. }
\label{sec:keyframe_extraction}
Efficient handling of data plays a vital role in the performance of video retrieval systems, particularly when dealing with large-scale video collections. Analyzing every single frame within a video is highly resource-consuming, resulting in unnecessary computational overhead, increased storage demand, and reduced retrieval efficiency. Therefore, we employed an optimized two-stage keyframe selection strategy, inspired by the TycheVid team \cite{10.1007/978-981-96-4291-5_14}. 

First, we utilize AutoShot \cite{zhu2023autoshotshortvideodataset} to segment the video into distinct shots, enabling us to focus on semantically consistent parts of the video. 

Second, within each shot, we apply an optimized keyframe selection algorithm. We use the \textbf{CLIP} model (ViT-L-14-quickgelu pre-trained with DFN2B) for semantic feature extraction. To optimize performance, features (embeddings) are extracted from every eighth frame within the shot. Let $\mathbf{e}_{\text{current}}$ be the embedding vector of the current sampled frame and $\mathbf{e}_{\text{prev}}$ be the vector of the last retained keyframe.

To quantify the visual change, we calculate the relative difference, $\text{rel\_diff}$, using the Euclidean ($L_2$) norm:
\begin{equation}
\label{eq:relative_difference}
\text{rel\_diff} = \frac{\Vert \mathbf{e}_{\text{current}} - \mathbf{e}_{\text{prev}} \Vert}{\Vert \mathbf{e}_{\text{prev}} \Vert}.
\end{equation}

A frame is retained as a new keyframe only if the computed $\text{rel\_diff}$ (Eq.~\ref{eq:relative_difference}) exceeds 0.4, a threshold empirically determined to yield effective results.

This two-stage strategy represents a deliberate trade-off between efficiency and granularity. The initial sampling of every eighth frame is a performance optimization to manage large-scale data. The adaptive component is the subsequent $L_2$-norm filtering. This filter ensures that only frames representing a significant visual change from this sampled set are retained, thus avoiding high redundancy in static scenes. While it is possible for a highly transient event occurring between the 8-frame intervals to be missed, this approach was calibrated to prioritize computational efficiency and robustly capture major scene changes, which proved highly effective for the event-level retrieval tasks in the competition.


\subsubsection{Multi-modal Feature Extraction. }
\label{sec:multi_modal_features}

Extracted keyframes are processed through a multimodal pipeline to generate metadata and embeddings. The Qwen2.5-VL-3B-Instruct model performs OCR and captioning, offering a strong trade-off between accuracy and efficiency~\cite{bai2025qwen25vltechnicalreport}. For embedding generation, we employ CLIP (DFN5B) for global semantic context and SigLIP2 for fine-grained recognition~\cite{radford2021learningtransferablevisualmodels,tschannen2025siglip2multilingualvisionlanguage}. These complementary models enhance retrieval flexibility across diverse query types.

\subsubsection{Automatic Speech Recognition. }
\label{sec:asr}





Audio features are extracted using Whisper~\cite{radford2022robustspeechrecognitionlargescale}, which outputs timestamped transcriptions. To align them with keyframes, each frame timestamp is matched to transcription intervals $(a, b)$ where $a \le t_k \le b$. The last spoken text is propagated through silent gaps, ensuring full coverage. The resulting transcription field enriches each keyframe’s metadata with synchronized speech content.

\subsection{Retrieval with Reciprocal Rank Fusion}
\label{sec:rrf}

\begin{figure}[t!]
\includegraphics[width=\textwidth]{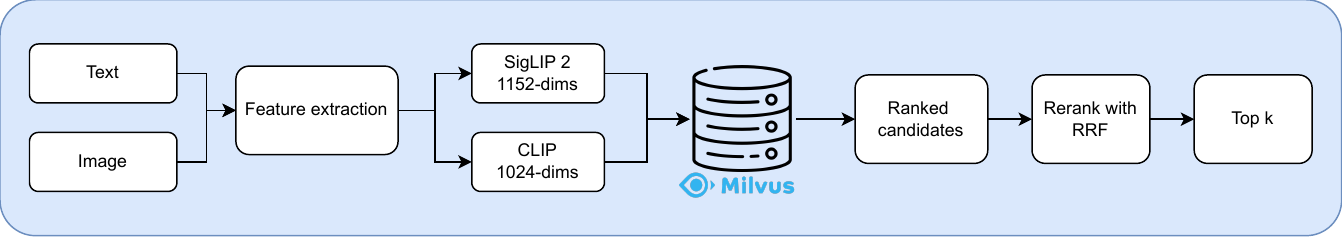}
\caption{Retrieval Module processes both text and image queries. SigLIP2 and CLIP query embeddings are stored in Milvus Retrieved candidates are then re-ranked using Reciprocal Rank Fusion (RRF), obtaining top-\emph{k} results.}
\label{fig:retrieval-module}
\end{figure}

Our system employs hybrid embeddings from CLIP and SigLIP2 for multimodal retrieval. Each user query, whether text or image, is embedded using both models, producing two independent similarity searches in Milvus and generating separate ranked lists of keyframes.

To merge these results into a single, more accurate ranking, we apply RRF~\cite{Cormack2009ReciprocalRF}, an effective and robust data-fusion method. For each keyframe $d$, the RRF score is computed as:
\begin{equation}
\label{eq:rrf_score}
\text{RRF\_Score}(d) = \sum_{i=1}^{N} \frac{1}{k + \text{rank}_i(d)},
\end{equation}
where $N=2$ (CLIP and SigLIP2) and $k$ is a constant (e.g., 60) that reduces sensitivity to lower-ranked results. Keyframes are then re-ranked by descending $\text{RRF\_Score}(d)$. This approach combines CLIP's global semantic understanding with SigLIP2's fine-grained recognition, substantially improving retrieval relevance and robustness.


\subsection{Query Refinement with Relevance Feedback}
\label{sec:relevance_feedback}
To enhance interactivity and support iterative refinement, we implemented a relevance feedback mechanism based on the Rocchio algorithm~\cite{10.1145/3366750.3366755}, a classical approach within the Vector Space Model. The algorithm updates the original query vector ($\vec{q_0}$) by shifting it toward the centroid of relevant keyframes and away from the centroid of non-relevant ones:
\begin{equation}
\label{eq:rocchio}
\vec{q_m} = \alpha \vec{q_0} + \beta \frac{1}{|C_r|} \sum_{\vec{d_j} \in C_r} \vec{d_j} - \gamma \frac{1}{|C_{nr}|} \sum_{\vec{d_j} \in C_{nr}} \vec{d_j},
\end{equation}
where $\alpha$, $\beta$, and $\gamma$ control the influence of the original query, relevant set ($C_r$), and non-relevant set ($C_{nr}$).

In our system's user interface, after an initial search, the user can provide explicit feedback by selecting "Prefer this answer" (like) or "Not prefer" (dislike) for any number of the returned images (keyframes). The system gathers all "liked" keyframes into the relevant set ($C_r$) and all "disliked" keyframes into the non-relevant set ($C_{nr}$).

This feedback is used to calculate the new query vector $\vec{q_m}$ using Equation~\ref{eq:rocchio}. This refined vector is then used to re-query the Milvus database, presenting the user with a new list of results that is more closely aligned with their stated preferences.

\subsection{Temporal Search}
\label{sec:temporal_search}
Standard retrieval systems often fail when handling complex temporal queries, such as those seen in the TRAKE challenge, as well as complex Known-Item Search (KIS) queries where the query itself defines a target frame based on its surrounding temporal context. These queries require the system to not only find individual events but also to verify that they occur in a specific sequence (e.g., an event $A$ happening before an event $B$) within the same video. A simple, single-stage search is insufficient for this task.

To address this, we developed a multi-stage temporal re-ranking algorithm (Algorithm~\ref{alg:temporal_reranking}). Our approach requires the user to decompose their query into three distinct components: a {current query} ($Q_{current}$), a {previous query} ($Q_{previous}$), and a {next query} ($Q_{next}$).

Our algorithm processes these inputs as three independent searches. First, the system retrieves three separate ranked lists of keyframes: $R_{current}$, $R_{previous}$, and $R_{next}$. The final ranking is then determined by re-scoring the main results in $R_{current}$.

For each current result $r_c$ (which belongs to a specific $\text{video\_ID}$) in the $R_{current}$ list, the algorithm searches the $R_{previous}$ and $R_{next}$ lists to find the highest-scoring keyframes, $r_p$ and $r_n$, that belong to the {exact same} $\text{video\_ID}$.

A new, temporally-boosted score ($S_{final}$) is then calculated for $r_c$ by summing its original score with the highest scores found for its corresponding "previous" and "next" keyframes (if any exist within that same video). This scoring can be formalized as:
\begin{equation}
\label{eq:temporal_score}
S_{final}(r_c) = S(r_c) + S_{\text{max}}(r_{p} \in \text{video\_ID}) + S_{\text{max}}(r_{n} \in \text{video\_ID}).
\end{equation}
If no matching previous or next keyframes are found in that video, their respective $S_{\text{max}}$ scores are treated as zero.

Finally, the $R_{current}$ list is re-sorted in descending order based on this new $S_{final}$. This re-ranking method ensures that a video containing the entire described sequence of events (Previous $\rightarrow$ Current $\rightarrow$ Next) will receive a significantly higher aggregate score, thus directly addressing the temporal alignment challenge and improving retrieval accuracy for complex sequential queries.

Here is the detailed complexity analysis of the multi-stage temporal re-ranking mechanism (Algorithm 1). Let $K$ be the number of top candidates retrieved from each of the three independent searches (for $Q_{previous}$, $Q_{current}$, and $Q_{next}$), which is a value set by the system.

\begin{itemize}
    \item \textbf{Step 1 (Independent Retrieval):} The Milvus HNSW index is employed to perform fast and high-recall Approximate Nearest Neighbor (ANN) retrieval, serving as the baseline computational cost for this step.

    \item \textbf{Step 2 (Compute Best Scores):} This step iterates through $R_{previous}$ ($K$ items) and $R_{next}$ ($K$ items). To ensure high performance, we implement \texttt{bestPrev} and \texttt{bestNext} as hash maps (or dictionaries). We build these maps by iterating through the two lists, a process with a time complexity of $O(K) + O(K) = O(K)$.

    \item \textbf{Step 3 (Temporal Re-scoring):} This step iterates through the main $R_{current}$ list ($K$ items). For each candidate $r_c$, retrieving its corresponding \texttt{bestPrev[video\_id]} and \texttt{bestNext[video\_id]} scores from the hash maps is an $O(1)$ operation on average. Therefore, the total complexity for this step is $O(K)$.

    \item \textbf{Step 4 (Re-ranking):} The final step sorts the re-scored list $R_{current}^{*}$, which has a size of $K$. This sorting operation has a time complexity of $O(K \log K)$.
\end{itemize}

\noindent
Thus, the total additional computational overhead of our entire temporal re-ranking algorithm (Steps 2-4) is dominated by the final sorting step, resulting in a complexity of $O(K \log K)$. This is a negligible cost compared to the initial ANN retrieval (Step 1) and confirms the algorithm's suitability for a real-time, interactive system.

The heuristic re-ranking approach is intentionally chosen over classical Dynamic Programming (DP) methods. Traditional DP algorithms are designed for dense alignment between two known, finite sequences and often incur substantial computational overhead, making them impractical to apply across the entire database in an interactive retrieval setting.

Our task, which includes both the Temporal Retrieval and Alignment of Key Events (TRAKE) and complex sequential KIS queries, requires finding a \emph{sparse sequence} of high-level semantic events ("Before", "Now", "After") within a massive video database.

Our method acts as a lightweight and effective heuristic. It leverages the speed of vector retrieval to find candidate events independently and then applies an $O(K \log K)$ re-ranking process to boost the scores of videos that contain the complete described sequence. This two-stage design - fast retrieval followed by efficient re-ranking - is far more scalable and better suited for an interactive retrieval system than a complex DP-based alignment that would be computationally infeasible to run against the entire database.

\begin{algorithm}[t!]
\caption{Multi-Stage Temporal Re-ranking for Sequential Queries}
\label{alg:temporal_reranking}
\begin{algorithmic}[1]
\Require Queries: $Q_{previous}$, $Q_{current}$, $Q_{next}$.
\Ensure Re-ranked list $R_{current}^{*}$ with temporally boosted scores.
\Statex

\State \textbf{Step 1: Independent Retrieval}
\State $R_{previous} \gets \text{Search}(Q_{previous})$
\State $R_{current} \gets \text{Search}(Q_{current})$
\State $R_{next} \gets \text{Search}(Q_{next})$

\Statex
\State \textbf{Step 2: Compute Best Scores per Video}
\ForAll{$(video\_id, score)$ in $R_{previous}$}
    \State $bestPrev[video\_id] \gets \max(bestPrev[video\_id], score)$
\EndFor
\ForAll{$(video\_id, score)$ in $R_{next}$}
    \State $bestNext[video\_id] \gets \max(bestNext[video\_id], score)$
\EndFor

\Statex
\State \textbf{Step 3: Temporal Re-scoring}
\ForAll{$(video\_id, frame\_id, S(r_c))$ in $R_{current}$}
    \State $S_{final}(r_c) \gets S(r_c) + bestPrev[video\_id] + bestNext[video\_id]$
    \State Append $(video\_id, frame\_id, S_{final}(r_c))$ to $R_{current}^{*}$
\EndFor

\Statex
\State \textbf{Step 4: Re-ranking and Output}
\State Sort $R_{current}^{*}$ in descending order of $S_{final}$
\State \Return $R_{current}^{*}$

\end{algorithmic}
\end{algorithm}

\begin{figure}[t!]
\centering
\includegraphics[width=\textwidth]{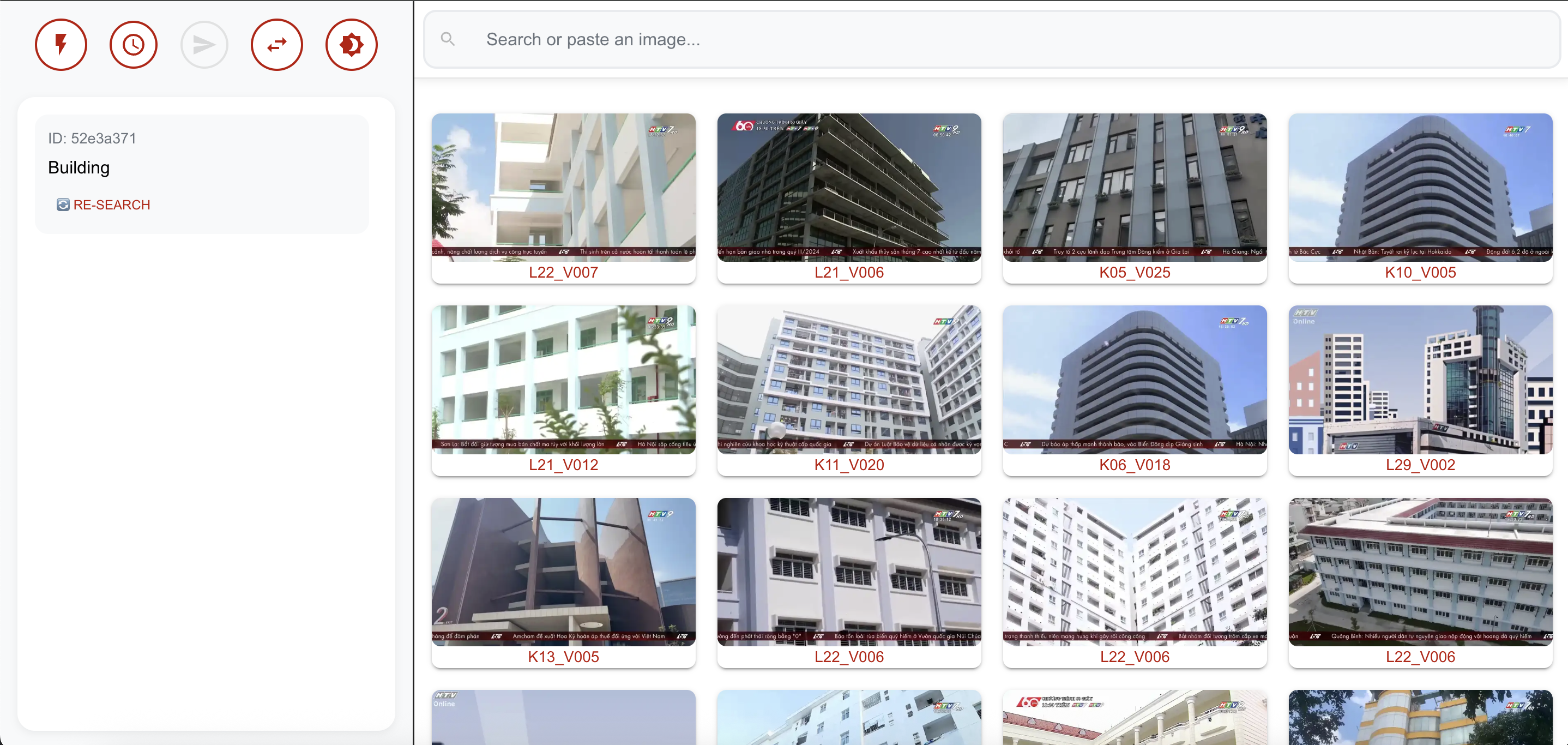}
\vspace{-5mm}
\caption{Overview of the system user interface. The left sidebar displays the query details and search management options, including the query ID and a re-search button. The right panel contains the search bar for text or image input at the top and presents the retrieved video keyframes as a grid of ranked results below.} 
\label{fig:UI}
\end{figure}

\begin{figure}[t!]
\centering
\includegraphics[width=\textwidth]{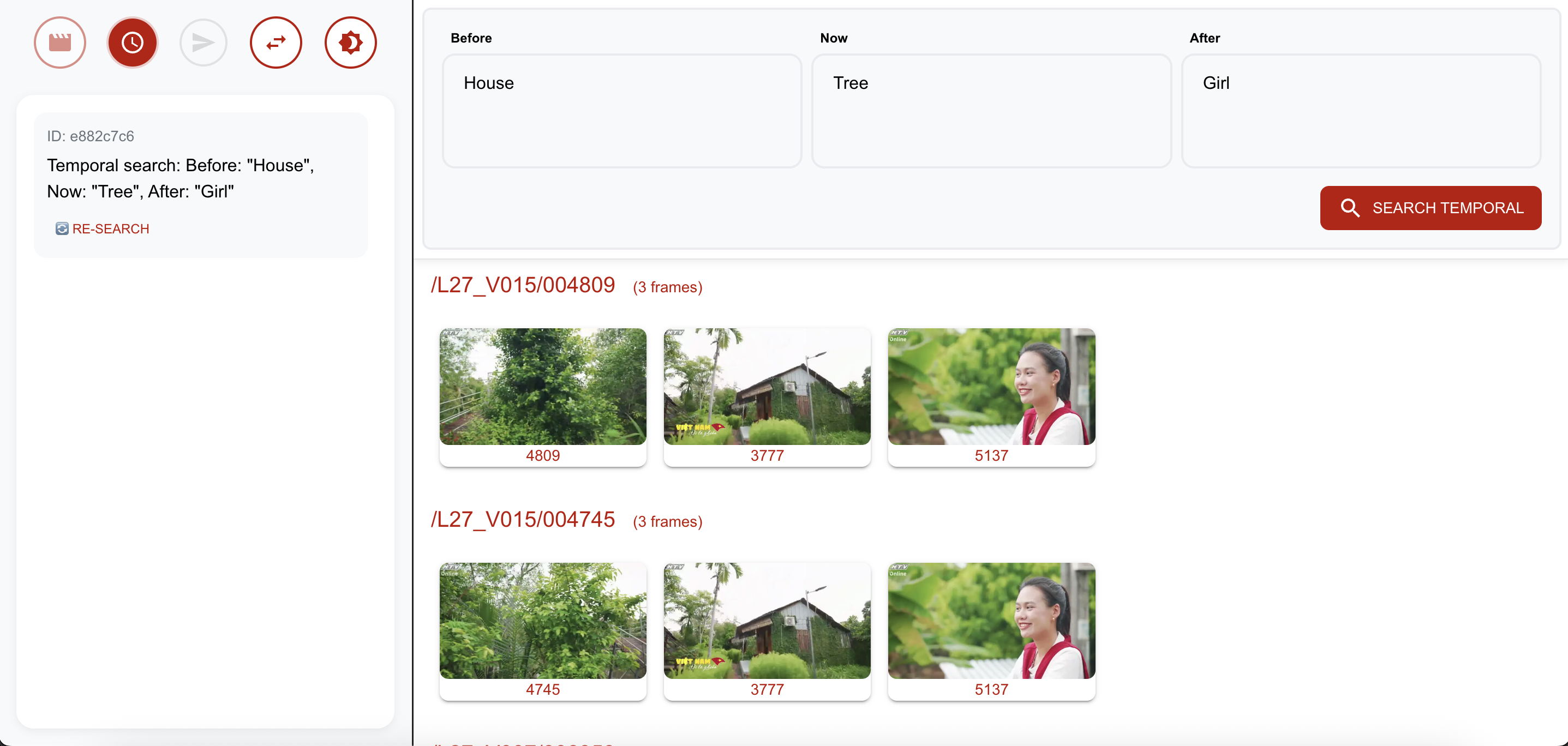}
\vspace{-5mm}
\caption{Temporal search mode interface, featuring three input fields for specifying temporal relations: Before, Now, and After.} 
\label{fig:temporal_mode}
\vspace{-3mm}
\end{figure}

\subsection{LLM-Assisted Query Interpretation}
While many contemporary retrieval systems employ Large Language Models (LLMs) for automatic query rewriting, this approach introduces risks such as intent drift and hallucination, where the rewritten query may diverge from the user's original goal. Given the real-time, interactive nature of the competition, maintaining user control and query fidelity is crucial.

To mitigate this, Vortex implements a hybrid, iterative refinement architecture. The LLM functions as a Query Interpretation Assistant rather than an autonomous rewriter. Instead of silently modifying the query, the system analyzes the user's input and proposes several explicit suggestions back to the user. This ensures the user retains full control and the query is not altered without their consent. This strategy is particularly effective for resolving semantic ambiguity. For instance, given a succinct or ambiguous query (e.g., ``building''), the LLM generates specific alternatives (e.g., ``a tall office building,'' ``a building under construction,'' ``a university campus building''). The user can then select the suggestion that best matches their intent, creating an iterative feedback loop that combines the LLM’s semantic power with explicit user guidance.

This design establishes a multi-layer refinement workflow. It combines LLM-assisted suggestions \textit{before} the search (pre-query refinement) with Rocchio-based Relevance Feedback \textit{after} the search (post-query refinement), providing a robust, transparent, and user-controlled retrieval experience.
\subsection{User interface}

To ensure efficient user interaction, we designed a clean and intuitive interface (Fig.~\ref{fig:UI}) with a left sidebar for navigation and a right panel for queries and visualization. The sidebar manages search sessions, history, and retrieval modes such as CLIP-only, SigLIP2-only, or the default RRF mode. The main panel provides a unified query bar for text and image inputs and displays ranked keyframes as search results. A dedicated Temporal Search Mode (Fig.~\ref{fig:temporal_mode}) supports complex sequential queries through three input fields labeled Before, Now, and After, allowing users to specify event order. This integrated design supports both standard and temporal search workflows seamlessly.



\section{Experiments}

\subsection{Dataset Overview}

Our system was evaluated on the official dataset of the AIC'25\footnote{\url{https://aichallenge.hochiminhcity.gov.vn/}}
. The dataset includes videos from major Vietnamese media channels such as 60 Giay Official, HTV Sports, Bao Thanh Nien, ViVu TV, HTV Giai Tri, HTV Entertainment, and Bao Tuoi Tre, covering diverse topics including news, sports, entertainment, and social events. In addition to the raw videos, the organizers provided keyframes, object annotations generated by a Faster R-CNN model pretrained on OpenImagesV4, CLIP (\texttt{ViT-B/32}) embeddings in \texttt{.npy} format, YouTube metadata (date, channel, and title), and a mapping file linking each frame ID to its timestamp and frame rate.


\subsection{Evaluation Metrics}

We evaluated the performance of Vortex using the official Mean of Top-k R-Score protocol defined by the AIC’25 organizers, which jointly measures retrieval accuracy and ranking quality. For each query, the system submitted up to 100 ranked results, with each result assigned an R-Score in the range [0, 1] according to its correctness against the ground truth. The scoring criteria varied by task: Textual-KIS required both the video name and frame index to fall within the reference range, Visual-QA required correct video, frame, and textual answer, and Temporal Alignment granted partial credit proportional to the number of correctly matched frames within the allowed tolerance.

For each query, the best R-Score within the top-$k$ results was computed at five cutoff levels ($k \in {1, 5, 20, 50, 100}$): $R@k = \max_{1 \leq i \leq k} R\text{-}Score(r_i)$. The Final Score for each query was defined as the mean of the five $R@k$ values, rewarding systems that retrieve correct answers early in the ranked list while maintaining consistent precision across different cutoff levels.

\subsection{Results}

\begin{table}[t!]
\centering
\caption{Official Final Scores of the proposed system across the three preliminary rounds of the AIC’25 competition.}
\label{tab:round_scores}
\begin{tabular}{lll}
\toprule
\textbf{Round} & \textbf{Final Score} & \textbf{Modules Integrated} \\
\midrule
Round 1 (24 queries) & 20.6 & Baseline (CLIP-only search) \\
Round 2 (30 queries) & 27.8 & Hybrid RRF (CLIP + SigLIP2) \\
Round 3 (35 queries) & 31.2 & Temporal Search + Relevance Feedback \\
\midrule
\textbf{Total} & \textbf{79.6/88 (90.5\%)} & Overall performance across all rounds \\
\bottomrule
\end{tabular}
\vspace{-3mm}
\end{table}


The performance of our system progressively improved as additional modules were integrated. The official scores for each competition round in the Preliminary Round are reported in Table~\ref{tab:round_scores}. Overall, the system achieved a final total of 79.6 out of 88 points, corresponding to approximately 90.5\% of the maximum possible score, demonstrating robust and consistent performance across all evaluation rounds.

The system’s performance in the Final Round was preliminarily evaluated by the Jury Board, and our Vortex system achieved an overall rating of \textit{Excellent}. The task-level evaluations are as follows: Excellent for TKIS, Very Good for VKIS, Very Good for TRAKE, and Outstanding for Q\&A. These results demonstrate the robustness of our system architecture in addressing the diverse and complex queries presented in the final round, and they highlight its particularly strong content-comprehension capabilities in the Q\&A task.

\subsection{System Usage Examples}

\begin{figure}[t!]
    \centering

    \begin{subfigure}{0.30\textwidth}
        \centering
        \includegraphics[width=\textwidth]{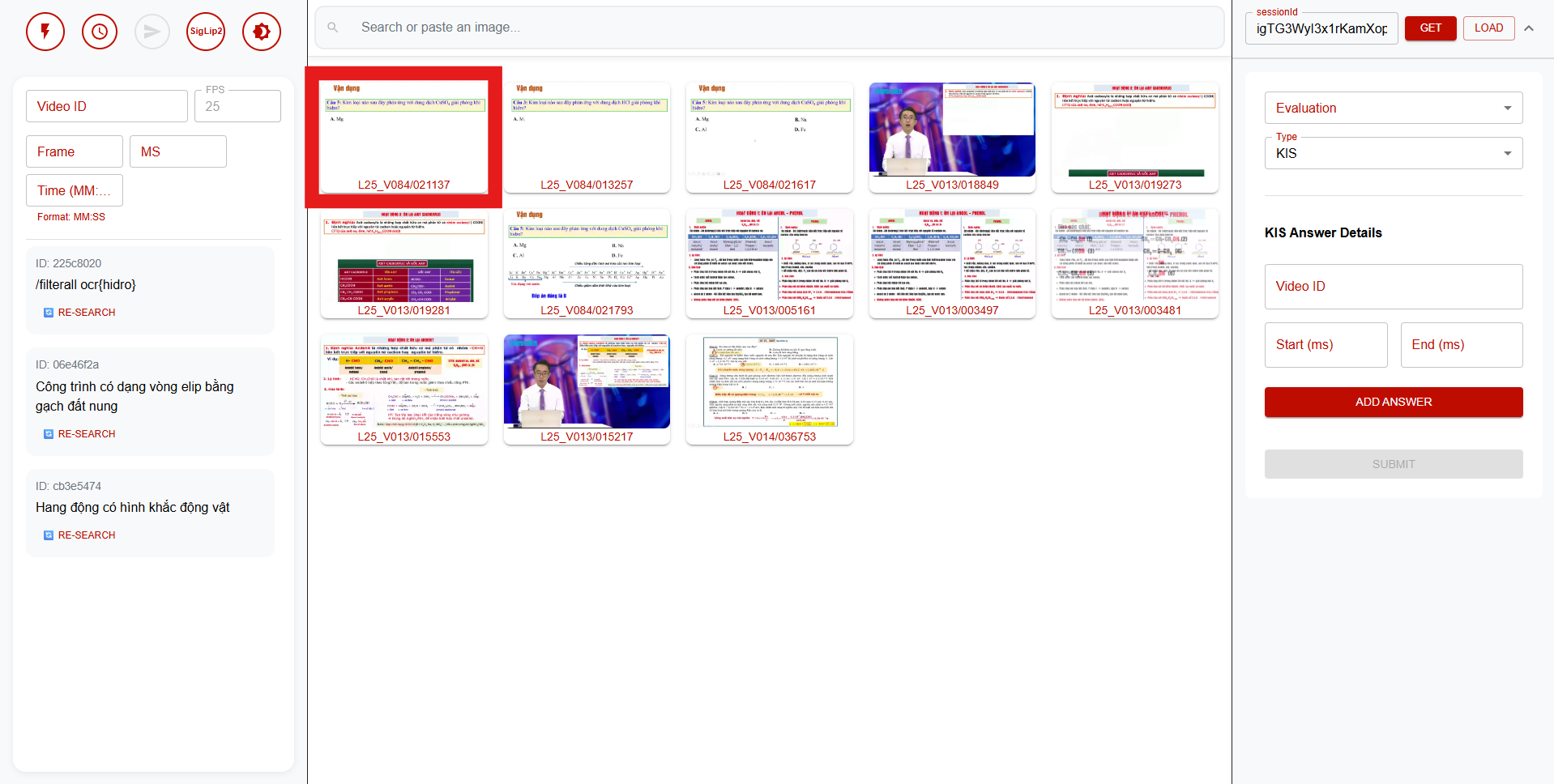}
        \caption{\textit{tkis-02}: Global OCR filter for "hidro".}
        \label{fig:tkis-2}
    \end{subfigure}
    \hfill
    \begin{subfigure}{0.30\textwidth}
        \centering
        \includegraphics[width=\textwidth]{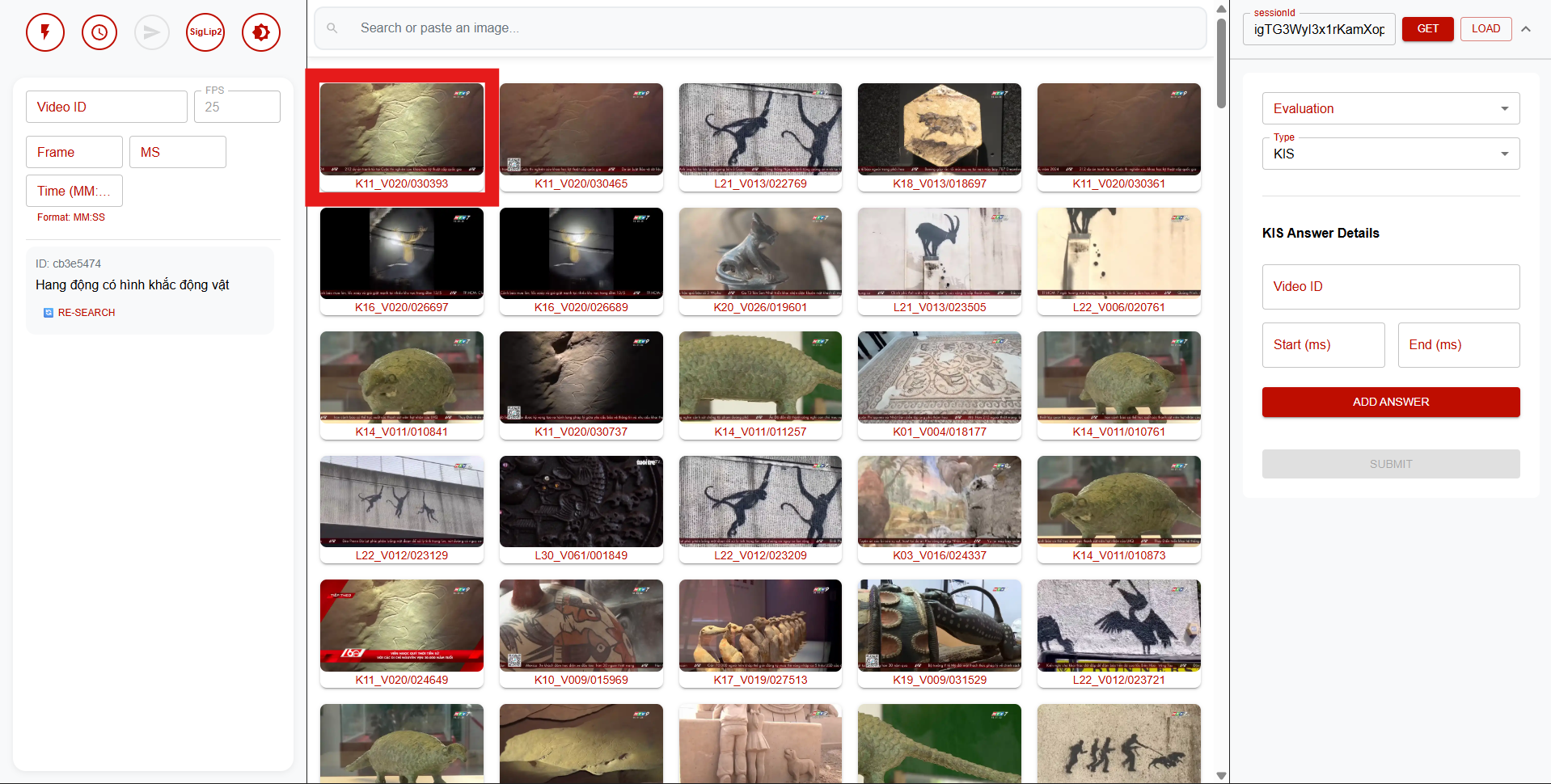}
        \caption{\textit{tkis-08}: Semantic search for "Hang dong...".}
        \label{fig:tkis-8}
    \end{subfigure}
    \hfill
    \begin{subfigure}{0.30\textwidth}
        \centering
        \includegraphics[width=\textwidth]{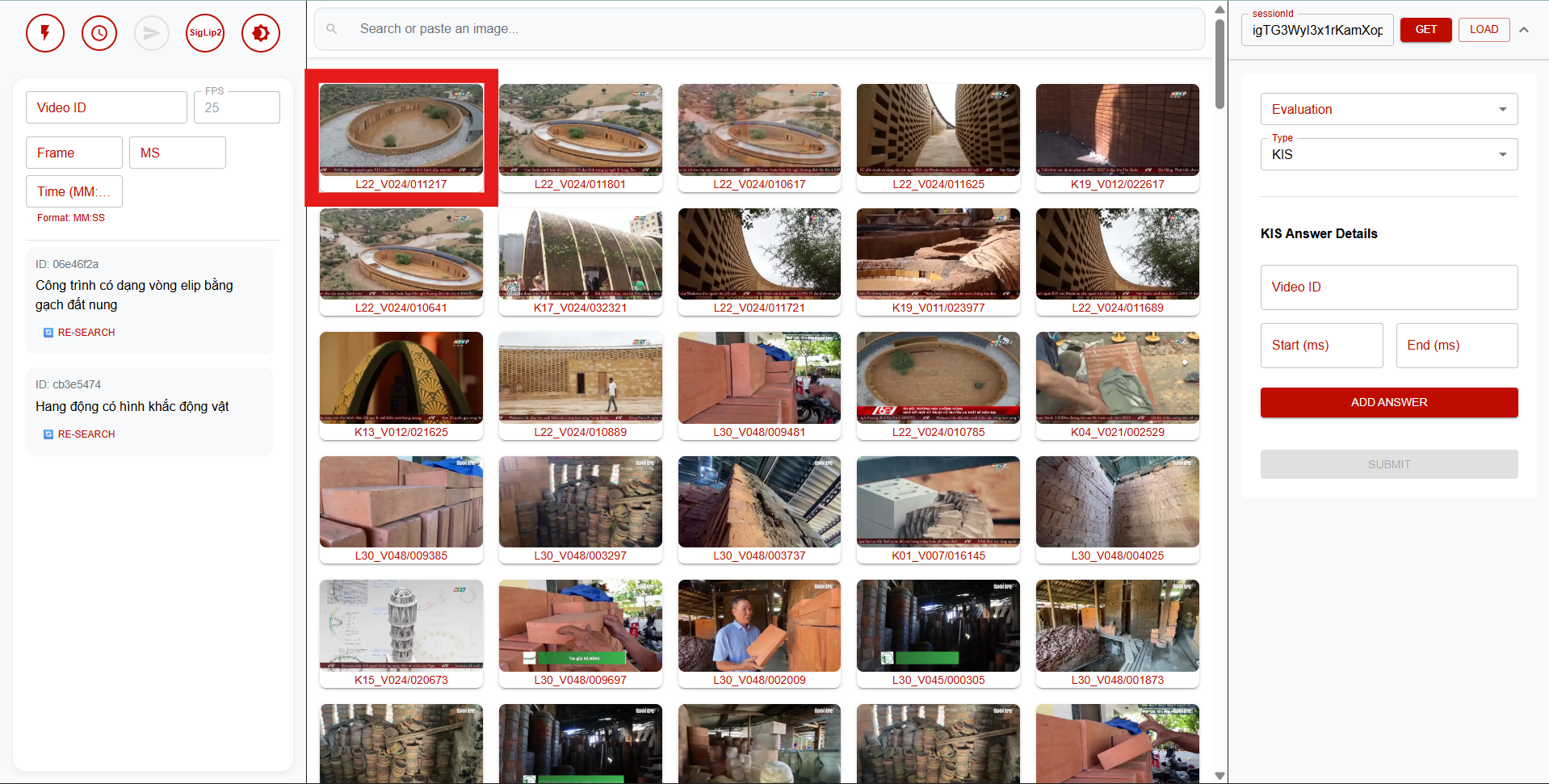}
        \caption{\textit{tkis-12}: Semantic search for "Cong trinh...".}
        \label{fig:tkis-12}
    \end{subfigure}

    \vspace{0.8em}

    \begin{subfigure}{0.30\textwidth}
        \centering
        \includegraphics[width=\textwidth]{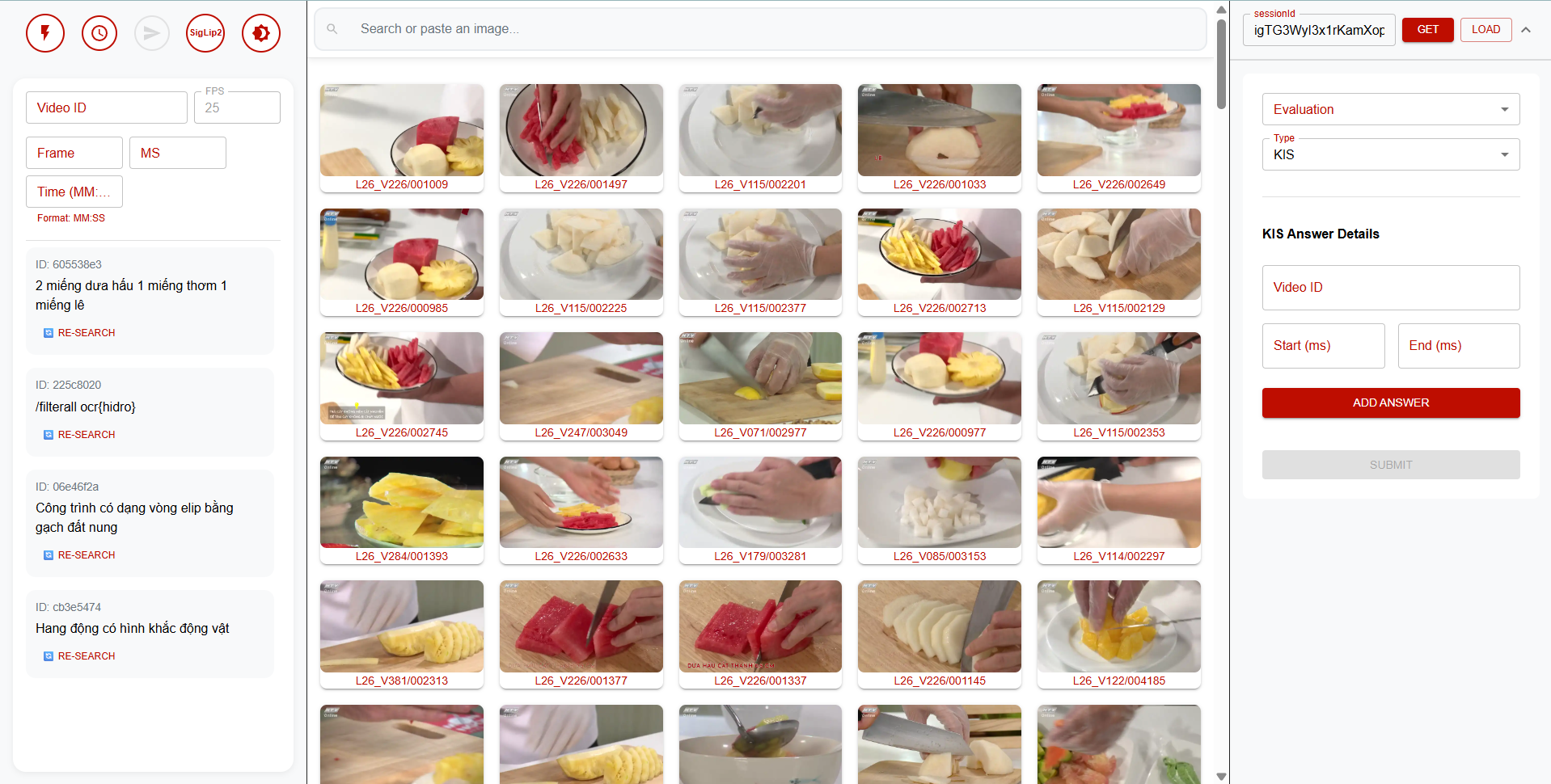}
        \caption{\textit{qa-02}: Initial search for ingredients.}
        \label{fig:qa-2-1}
    \end{subfigure}
    \hfill
    \begin{subfigure}{0.30\textwidth}
        \centering
        \includegraphics[width=\textwidth]{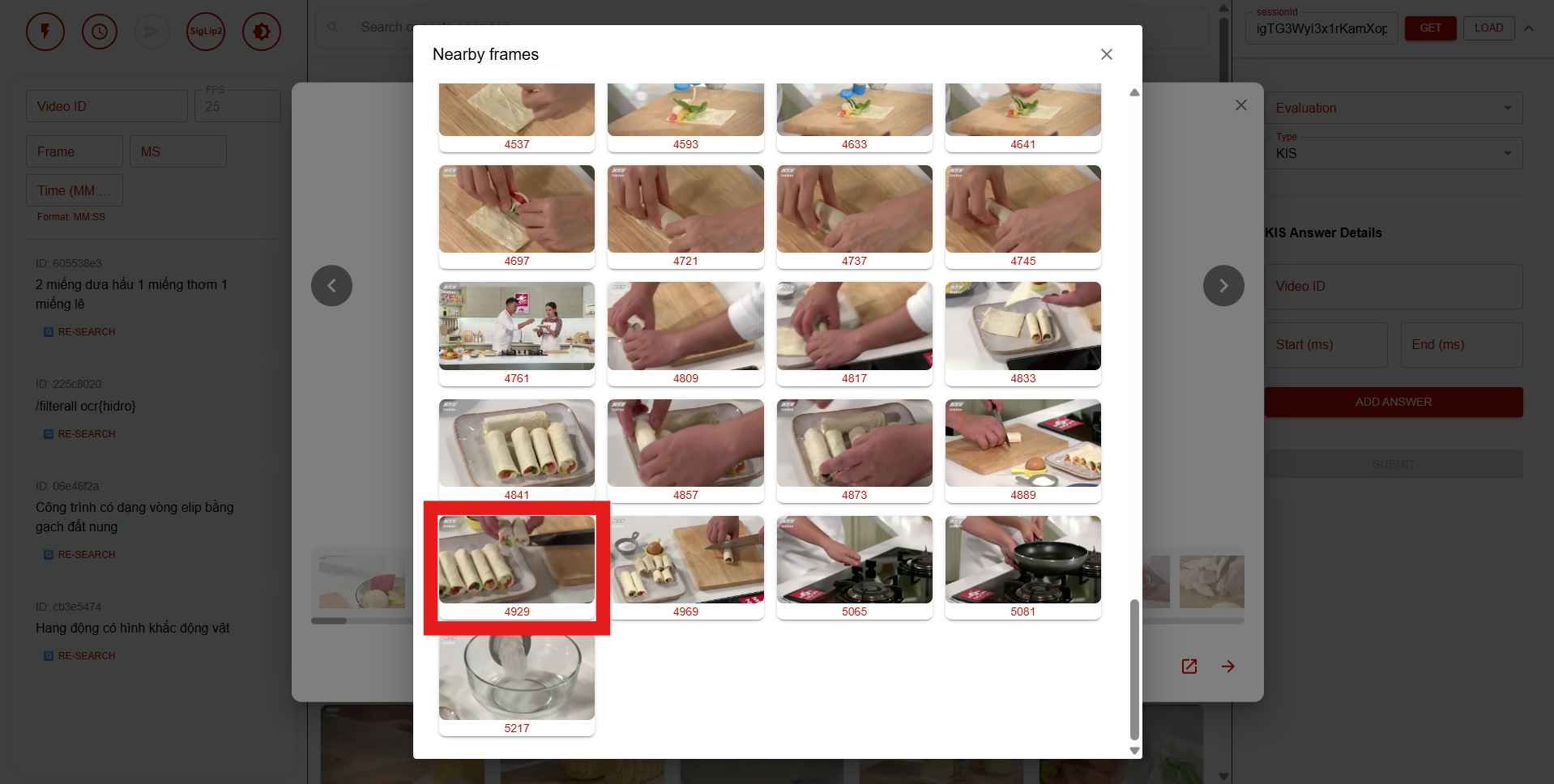}
        \caption{\textit{qa-02}: "Nearby frame" feature showing the cut.}
        \label{fig:qa-2-2}
    \end{subfigure}
    \hfill
    \begin{subfigure}{0.30\textwidth}
        \centering
        \includegraphics[width=\textwidth]{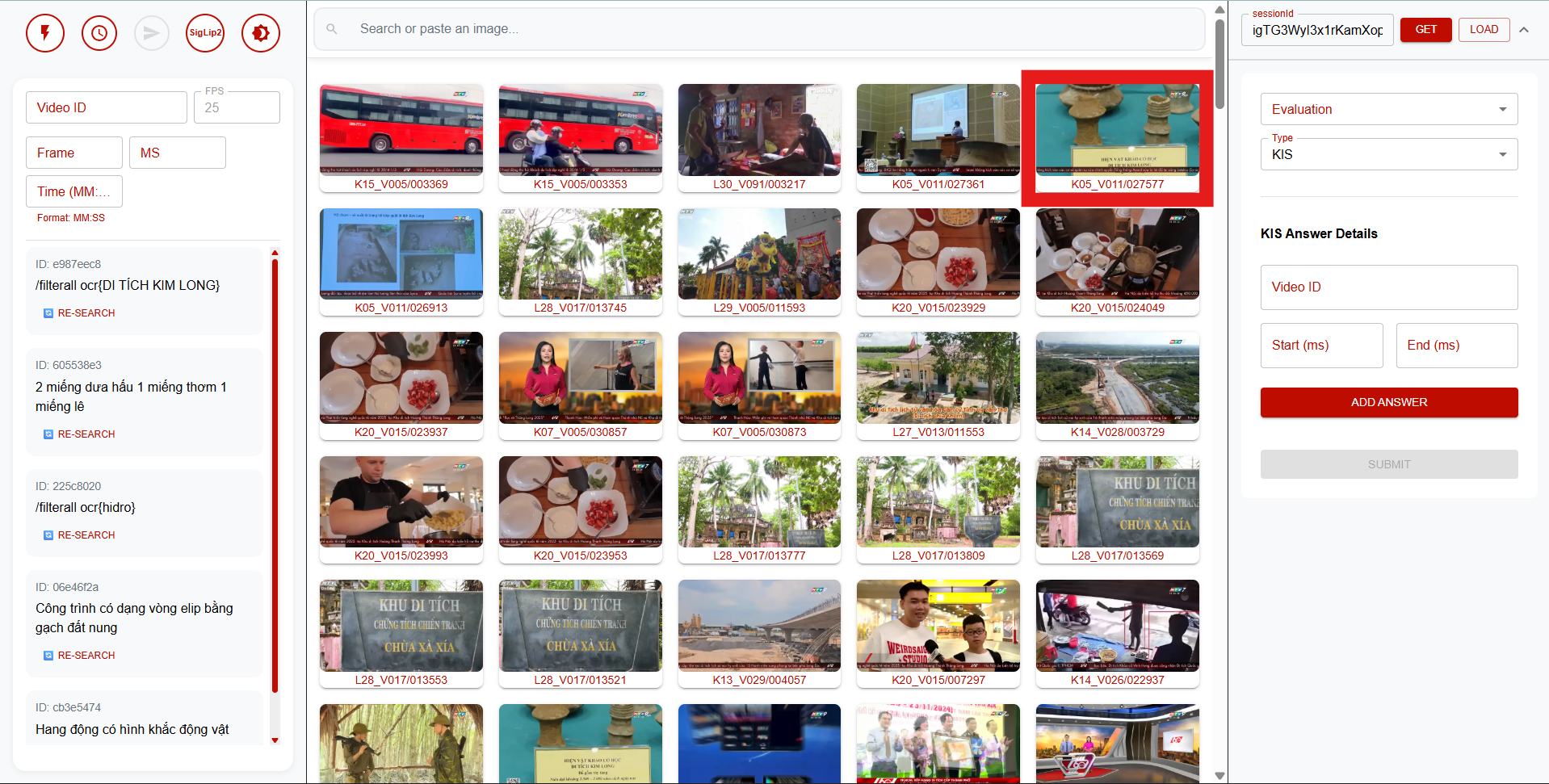}
        \caption{\textit{vkis-07}: OCR filter for "DI TICH KIM LONG".}
        \label{fig:vkis-7}
    \end{subfigure}

    \vspace{0.8em}

    \begin{subfigure}{0.30\textwidth}
        \centering
        \includegraphics[width=\textwidth]{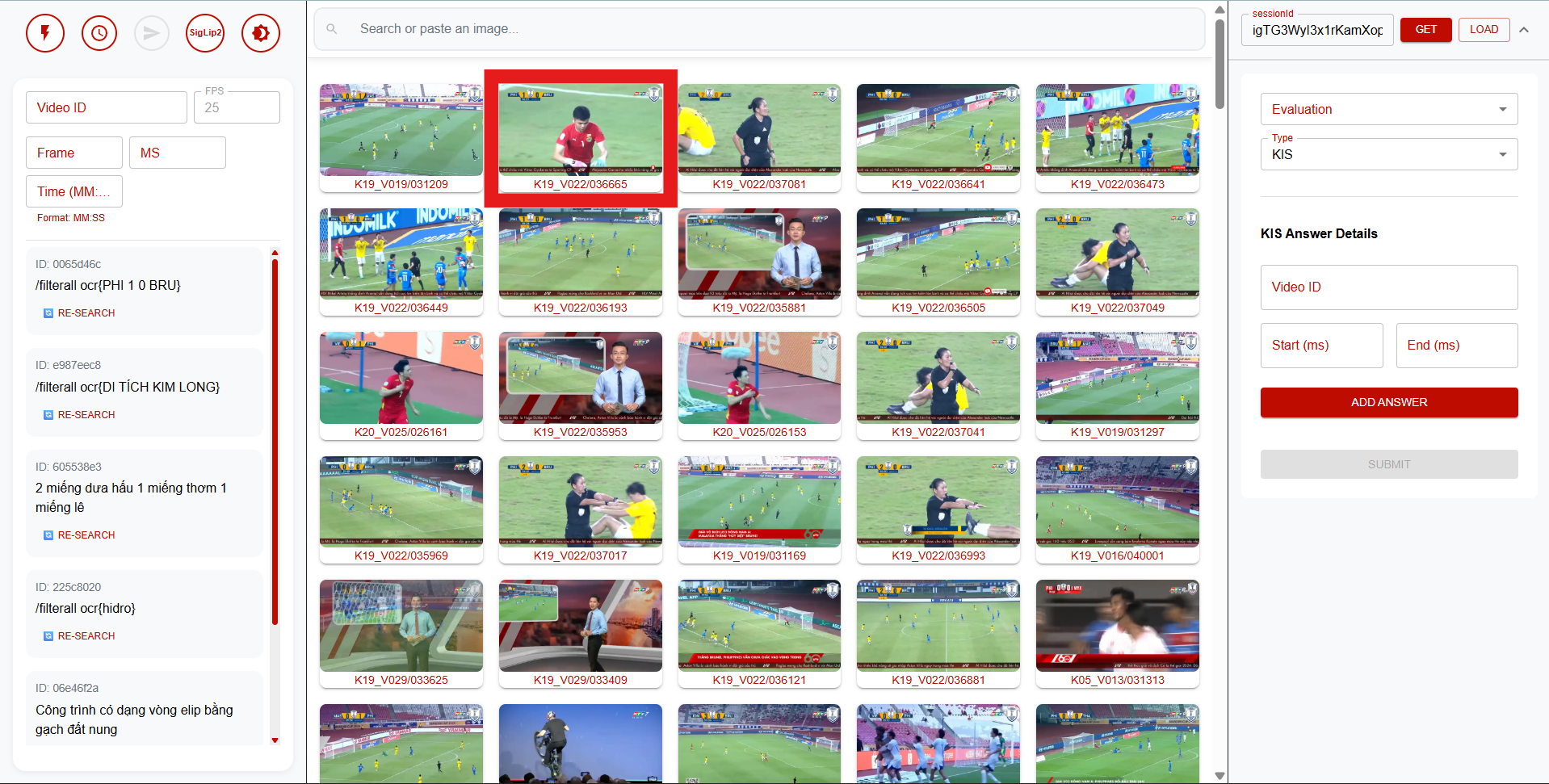}
        \caption{\textit{trake-03} (E1, Step 1): OCR search for "PHI 1 0 BRU" finds a frame `after' the goal.}
        \label{fig:trake-e1-1}
    \end{subfigure}
    \hfill
    \begin{subfigure}{0.30\textwidth}
        \centering
        \includegraphics[width=\textwidth]{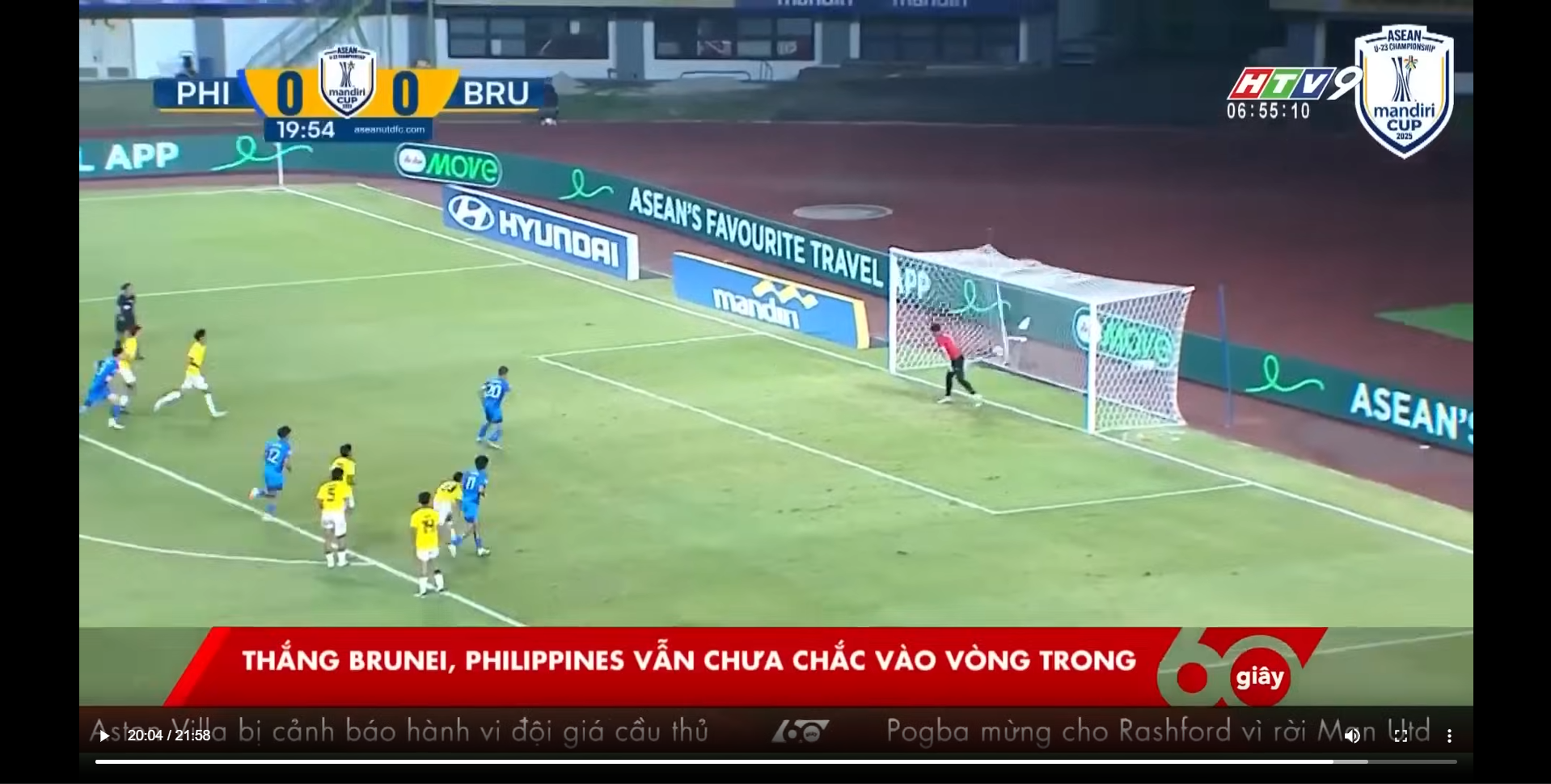}
        \caption{\textit{trake-03} (E1, Step 2): Exploring nearby frames prior to the OCR result to find the event sequence.}
        \label{fig:trake-e1-2}
    \end{subfigure}
    \hfill
    \begin{subfigure}{0.30\textwidth}
        \centering
        \includegraphics[width=\textwidth]{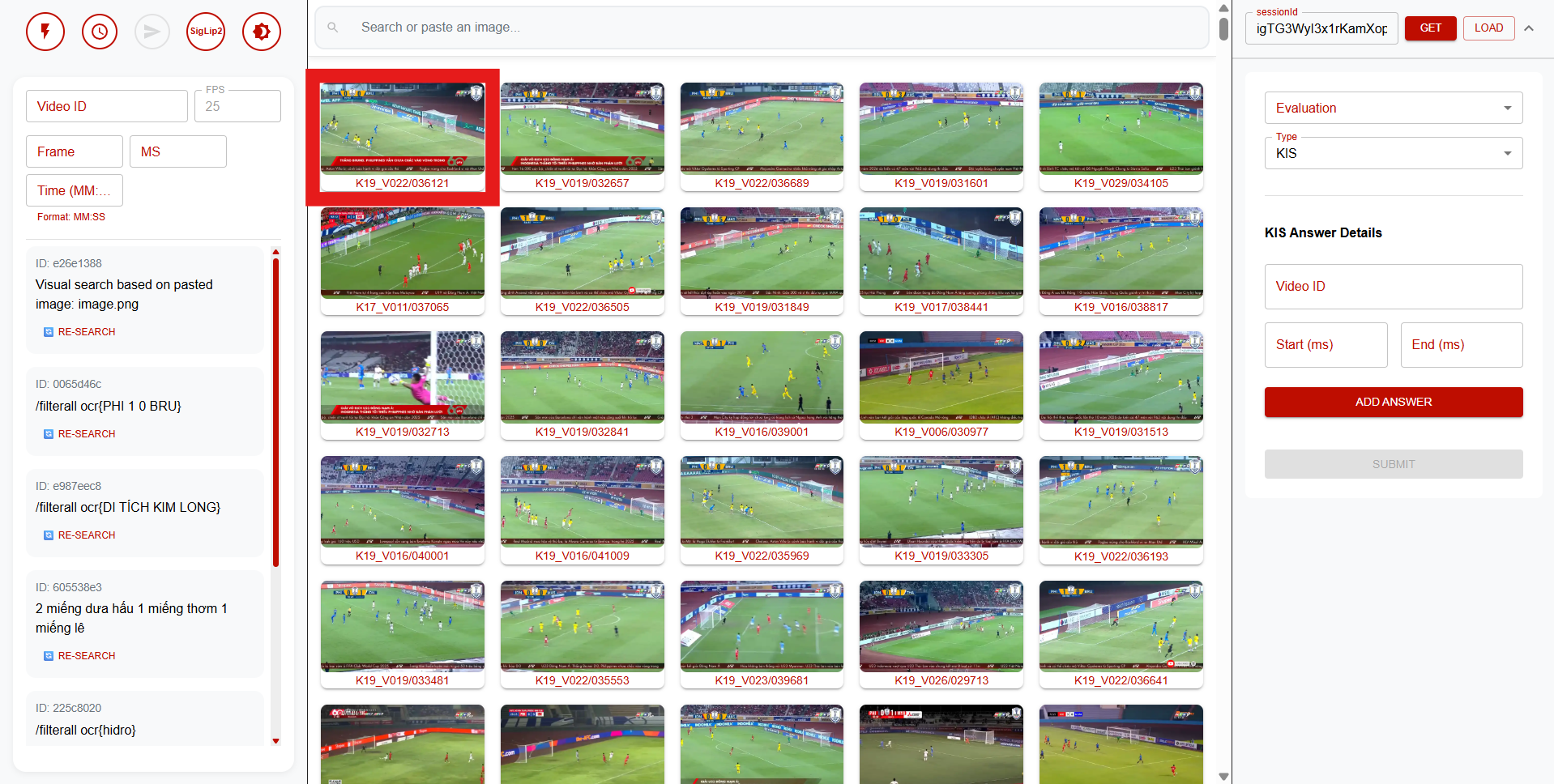}
        \caption{\textit{trake-03} (E1, Step 3): Visual Search pinpoints the exact keyframe of the goal.}
        \label{fig:trake-e1-3}
    \end{subfigure}

    \caption{Screenshots of the Vortex system in action for various query types.}
    \label{fig:system_usage}
    \vspace{-3mm}
\end{figure}

To demonstrate the practicality and interactivity of the Vortex system, we illustrate strategies used for several representative competition queries (Fig.~\ref{fig:system_usage}):

\textbf{Query tkis-query-02 (Textual KIS):}
The hint referenced a multiple-choice question about a reaction that “releases hydrogen gas” (non-diacritic: ``giai phong khi hidro''). Because this was a highly specific cue, we skipped semantic search and directly applied the global OCR filter: \texttt{/filterall ocr\{hidro\}}. This immediately surfaced the correct video (Fig.~\ref{fig:tkis-2}).

\textbf{Query tkis-query-08 (Textual KIS):}
The hint described “a cave in the country famous for Gaulois roosters, with animal and human engravings.” We issued the semantic query ``Hang dong co hinh khac dong vat'' (Eng: “Cave with animal engravings”), and the system retrieved relevant clips via ASR transcripts and generated captions (Fig.~\ref{fig:tkis-8}).

\textbf{Query tkis-query-12 (Textual KIS):}
The hint mentioned “an elliptical structure made of baked bricks.” Using the semantic query ``Cong trinh co dang vong elip bang gach dat nung,'' the system returned the correct segment based on scene descriptions (Fig.~\ref{fig:tkis-12}).

\textbf{Query qa-query-02 (Q\&A):}
The question asked how many pieces a sandwich was cut into. The first hint listed the ingredients (``2 strips of watermelon, 1 strip of pineapple, 1 strip of pear''). Searching this phrase returned the clip’s starting point (Fig.~\ref{fig:qa-2-1}). Using the ``nearby frame'' feature, we quickly navigated to the cutting step, confirming the answer is two (Fig.~\ref{fig:qa-2-2}).

\textbf{Query vkis-07 (Video KIS):}
This query provided no textual hints. By observing the displayed video, we identified the phrase ``DI TICH KIM LONG'' (Eng: “KIM LONG HISTORICAL SITE”) on an artifact and used the OCR filter to search it. The system immediately located the correct footage (Fig.~\ref{fig:vkis-7}).

\textbf{Query trake-03 (TRAKE):} This query required identifying three events: {E1} (first goal), {E2} (penalty save), and {E3} (second goal). Our coarse-to-fine approach for \textbf{E1} illustrates the system’s interactivity:
\begin{itemize}
    \item \textbf{Step 1 (Coarse Filter):} We applied an OCR filter for the scoreboard update ``PHI 1 0 BRU,'' which returned a frame immediately \emph{after} the goal (Fig.~\ref{fig:trake-e1-1}).
    \item \textbf{Step 2 (Temporal Search):} Using the ``nearby frame'' tool, we navigated backward to inspect the sequence leading up to the goal (Fig.~\ref{fig:trake-e1-2}).
    \item \textbf{Step 3 (Fine-Grained Search):} We then captured a frame from this sequence and performed a Query-by-Example visual search, pinpointing the exact moment the ball crossed the line (Fig.~\ref{fig:trake-e1-3}).
\end{itemize}

This multi-stage strategy, such as text filtering, temporal navigation, and visual refinement, was also used to locate events {E2} and {E3}.

\section{Conclusion}

This paper introduced Vortex, a unified multi-modal video retrieval system developed for the AIC'25, which integrates semantic understanding, fine-grained visual cues, and interactive query refinement within a single framework. In the competition, Vortex delivered strong and consistent results, scoring 79.6/88 (90.5\%) in the Preliminary Round and earning an "Excellent" overall rating in the Final Round. Notably, it achieved "Outstanding" performance in Q\&A and excelled across the remaining tasks. These outcomes highlight the effectiveness of combining semantic retrieval with fine-grained content analysis, as well as the value of interactive search mechanisms. Vortex demonstrates a powerful direction for context-aware, adaptable, and user-centered video retrieval systems, providing a solid foundation for future extensions in large-scale and real-world multimedia search.




\begin{credits}
\subsubsection{\ackname} 
This research is supported by research funding from Faculty of Information Technology, University of Science, Vietnam National University - Ho Chi Minh City.
\end{credits}

%
%
%
%

\bibliographystyle{splncs04}
\bibliography{references}







\end{document}